\documentclass[10pt,journal]{IEEEtran} 

\ifCLASSOPTIONcompsoc
  \usepackage[nocompress]{cite}
\else
  \usepackage{cite}
\fi

\usepackage{lscape}
\usepackage{etex}
\usepackage{graphicx}
\usepackage{multirow}
\usepackage{amsmath}
\usepackage{amssymb}
\usepackage{color}
\usepackage{tikz}
\usepackage{url}
\usepackage{placeins}
\usepackage{float}
\usepackage{multirow}
\usepackage{array}
\usepackage{footmisc}
\usepackage{wrapfig}
\usepackage[bf,sc,small]{caption} 
\usepackage{listings}
\usepackage{algorithm, algpseudocode}
\algnewcommand{\Inputs}[1]{%
	\State \textbf{Inputs:}
	\Statex \hspace*{\algorithmicindent}\parbox[t]{.8\linewidth}{\raggedright #1}
}
\algnewcommand{\Output}[1]{%
	\State \textbf{Output:}
	\Statex \hspace*{\algorithmicindent}\parbox[t]{.8\linewidth}{\raggedright #1}
}
\algnewcommand{\Initialize}[1]{%
	\State \textbf{Initialize:}
	\Statex \hspace*{\algorithmicindent}\parbox[t]{.8\linewidth}{\raggedright #1}
}

\newcommand\numberthis{\addtocounter{equation}{1}\tag{\theequation}}
\usepackage{layouts}
\usepackage{tabularx}
\usepackage{mathtools}
\usepackage{adjustbox}
\usepackage{xcolor}
\usepackage{enumitem}

\usepackage[caption=false]{subfig}
\usepackage{xspace}
\usepackage{changepage}
\usepackage{epsfig}
\usepackage{booktabs}
\usepackage{hyperref}
\usepackage{microtype}  	

\hypersetup{colorlinks=true, urlcolor=black, citecolor=blue, linkcolor = black}

\DeclareMathAlphabet{\mathitbf}{OML}{cmm}{b}{it}
\DeclareMathAlphabet{\mathbfit}{OML}{cmm}{b}{it}

\definecolor{mygreen}{rgb}{0,0.6,0}
\definecolor{myred}{rgb}{1,0.333,0.333}

\usepackage{amsmath}
\usepackage{amsthm}
\usepackage{amssymb}
\usepackage{tikz}
\usepackage[all,cmtip]{xy}
\theoremstyle{definition}

\newtheorem{lem}{Lemma}

\newcommand\autorefeq[1]{\hyperref[#1]{Equation~\eqref{#1}}}%
\newcommand\autorefapp[1]{\hyperref[#1]{Appendix~\ref{#1}}}%
\newcommand\autorefalg[1]{\hyperref[#1]{Algorithm~\ref{#1}}}%
\newcommand\autorefcor[1]{\hyperref[#1]{Corollary~\ref{#1}}}%
\newcommand\autorefprop[1]{\hyperref[#1]{Proposition~\ref{#1}}}%
\newcommand\autorefproperty[1]{\hyperref[#1]{Property~\ref{#1}}}%
\usepackage[utf8]{inputenc}
\usepackage{pict2e,picture}
\usepackage[english]{babel}

\newsavebox\CBox
\newlength\CLength
\def\Circled#1{\sbox\CBox{#1}%
  \ifdim\wd\CBox>\ht\CBox \CLength=\wd\CBox\else\CLength=\ht\CBox\fi
    \makebox[1.2\CLength]{\makebox(0,1.2\CLength){\put(0,0){\circle{1.4\CLength}}}%
    \makebox(0,1.2\CLength){\put(-.5\wd\CBox,0){#1}}}}

\newif\ifanonymous

\begin{document}

\title{A Search and Detection Autonomous Drone System: from Design to Implementation}

\ifanonymous
\author{}
\institute{}
\else
\author{Mohammadjavad Khosravi,
	Rushiv Arora,
	Saeede Enayati, and 
	Hossein Pishro-Nik, \textit{Member, IEEE} 
	\IEEEcompsocitemizethanks{
		\IEEEcompsocthanksitem M. Khosravi, R. Arora, S. Enayati, and H. Pishro-Nik are with the Department of Electrical and Computer Engineering, University of Massachusetts, Amherst, MA, 01003 USA\protect, E-mail: mkhosravi@umass.edu, rrarora@umass.edu, senayati@umass.edu, pishro@engin.umass.edu.}
	\thanks{This work was supported by NSF under grant CNS-1932326.}}

\fi
\renewcommand\footnotemark{}

\maketitle



\begin{abstract}
	Utilizing autonomous drones or unmanned aerial vehicles (UAVs) has shown great advantages over preceding methods in support of urgent scenarios such as search and rescue (SAR) and wildfire detection. In these operations, search efficiency in terms of the amount of time spent to find the target is crucial since with the  passing of time the survivability of the missing person decreases or wildfire management becomes more difficult with disastrous consequences. In this work, it is considered a scenario where a drone is intended to search and detect a missing person (e.g., a hiker or a mountaineer) or a potential fire spot in a given area. In order to obtain the shortest path to the target, a general framework is provided to model the problem of target detection when the target's location is probabilistically known. To this end, two algorithms are proposed: Path planning and target detection. The path planning algorithm is based on Bayesian inference and the target detection is accomplished by means of a residual neural network (ResNet) trained on the image dataset captured by the drone as well as existing pictures and datasets on the web. Through simulation and experiment, the proposed path planning algorithm is compared with two benchmark algorithms. It is shown that the proposed algorithm significantly decreases the average time of the mission. 
	

\end{abstract}


\IEEEpeerreviewmaketitle

\section{Introduction}\label{sec:intro}

Autonomous drones or unmanned aerial vehicles (UAVs) represent promising advantages over conventional human-intervened target detection methods due to their fast deployment, autonomous mobility capability, low cost, and easier reachability to the hard-to-reach areas. In a typical scenario, UAVs will be deployed in an area of interest, perform sensory operations to collect evidence of the presence of a target, and report their collected information to a remote ground station or a rescue team \cite{adams2007search,human2015}.

 \textcolor{black}{However, to detect objects in these scenarios using learning algorithms, especially neural networks, large and relevant datasets are required \cite{mishra2020drone}. On the other hand, to decrease the detection and rescue operation's delay, it is essential to design time-efficient path planning algorithms for the UAVs which is not as straightforward as other operations such as object tracking, package delivery, imaging, etc., as the object's location is not known, or it is stochastically known. To improve the efficiency of using UAVs in such situations, multiple UAVs operations have been considered as well since multiple UAVs can significantly increase the speed of the operation or equivalently decrease its time duration which is highly desirable. However, designing an optimal coordination among the UAVs such that no overlap exists in their monitoring regions, as well as cost increase are the challenges of multiple UAV operations.}

In this paper, scenarios where an autonomous drone is deployed in a region to accomplish a target detection task are considered with a focus on two examples of such systems: (1) A UAV is used to search for a lost person (e.g., a mountain climber lost due to avalanche, snow, etc.) who could be in a dire situation and might need immediate help. This scenario is referred to as the search and rescue operations (SAR) scenario. (2) The UAV is deployed in a forest area in surveillance mode to detect a potential fire spot \footnote{In this paper, the fire spot is assumed to be in its initial states. Therefore, it is treated as a point object and no propagation model has been considered for the fire.}. This scenario is referred to as the fire surveillance. Both applications are of great interest\cite{khan2019exploratory,aydin2019use,lum2015automatic,rashid2020compdrone,park2020wildfire,viseras2019wildfire,pan2020computationally,schedl2021autonomous,karaca2018potential,hayat2020multi,dominguez2017planning,eyerman2018drone,pereira2021optimal,dousai2021detection}.

An autonomous drone searching for a lost person or a potential fire spot makes a high-impact decision in its next move which will have two possible outcomes: (1) it might choose a correct path and gets closer to the object and (2) it might choose a wrong path and gets further away from the object. The former leads to finding the object in a shorter time while the latter can result in a significant delay. Hence, it is crucial that the target is found within the shortest amount of time, as any delays in finding the target results in the decrease of the survivability of the missing person or the wildfire management becomes more difficult. Thus, the most important goal is to minimize detection time.

\begin{figure}[htb]	
	\centering
	\includegraphics[scale = 0.4]{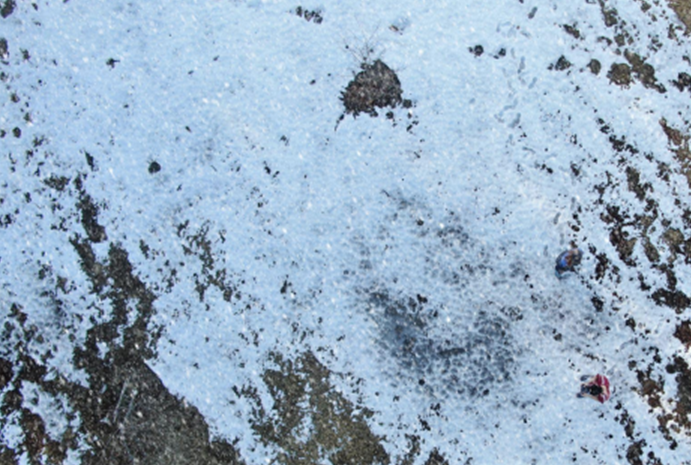}
	\caption{An example of an aerial image from a snowy mountain with a person at the bottom right corner. }
	\label{fig.SARcomplexity}
\end{figure}

In these operations, the drone uses its camera to monitor the area and find the target. Since the target objects are relatively small and often camouflaged within the environment, it is very important that a detection algorithm can accurately recognize the target from its surroundings. A typical image of aerial images of a snow-covered area is shown in Figure \ref{fig.SARcomplexity} where a person is also located at the bottom right corner of the picture.

In this paper, a path planning algorithm for a drone is developed to find the shortest path to the target while the location of the target is not deterministically known. Using a prior knowledge of the target location's distribution, the path planing algorithm covers the whole area in order to detect the target, while the drone will follow the chosen path and collect data for processing.
A Machine learning (ML) technique is implemented to quickly and accurately detect the target. Specifically, a residual neural network (ResNet) is utilized in which the dataset used for the training and validation is a combination of the existing datasets as well as pictures taken by the drone where in case of fire surveillance, mostly include small fire spots or smoke in a rural area with other objects around (spare). This is because the goal is to achieve an early detection where the fire is in its initial stages. 
 
 The contributions of this paper are as follows:
 
 \begin{itemize}
 	\item A general framework is developed to model the problem of finding a target through the shortest path when the location is probabilistically known. In this regard, first a Bayesian framework is proposed to estimate the target location. Then, using the sliding windowing technique for the path planning problem a suboptimal solution is obtained.
 	
 	\item Using ResNet, a target detection algorithm is designed with an accuracy of greater than $91\%$ for the training and validation pictures taken from the existing datasets and images generated by the drone. 
 	
 	\item Through simulation, the proposed path planning algorithm is compared with two benchmark algorithms and it is shown that the proposed algorithm significantly decreases the average time of the mission and the energy consumption of the drone in the two types of probability maps considered.  
 	
 	\item Through experimentations, the proposed framework is applied to a fire detection scenario where it is shown that the proposed algorithm outperforms the benchmark algorithms in terms of time spent to find the target and the energy consumption of the drone.

 \end{itemize}

The rest of the paper is organized as follows: in Section II, related works are provided. In Section III, a simplified scenario, a Naive algorithm, and the proposed system model are presented. In Section IV, the proposed algorithm and machine learning architecture are provided. Section V provides the simulation results, and Section VI concludes the paper.

\section{Related Work}
UAV-based target detection and target tracking has been increasingly attracting attentions in the civil applications \cite{Minaeian2018,wang2019autonomous,wang2020,wang2020social,huang2021,chen2021high}. 
Meanwhile, there have been numerous proposals in utilizing UAVs in SAR operations in recent years, from post-natural or human-made disasters such as earthquakes and explosions to offshore SAR operations in oil-rigs platforms and monitoring wildfire spread rate \cite{hazim2019unmanned,eid2010design,lin2019kalman}.

SAR and wildfire detection operations can be investigated from different points of view, e.g., object detection, optimal routing, optimal resource allocation, etc. Generally speaking, majority of works in these scenarios have allotted to develop efficient and accurate object detection algorithms, for example see \cite{cui2015drones,viseras2019,quan2019project,wildfire2019hossain,Domozi2020,Mishra2020DronesurveillanceFS,sasa2021} where different image datasets were used to train the learning algorithms. 

As far as target detection is of concern, using UAVs in wildfire detection has attracted a lot of attention as well in recent years \cite{yuan2016,Yuan2017,smoke2016,alireza2020transfer}. In \cite{yuan2016}, a detection method
utilizing both color and motion features is investigated for UAV-based fire detection. The same authors in \cite{Yuan2017} developed a detection algorithm using Infrared images in order to detect fire pixels. 
 Using smoke sensors, a smoke detection algorithm was proposed in \cite{smoke2016}. In \cite{alireza2020transfer}, in order to use a smaller dataset and reduce the computational
complexity, the authors proposed to use a pretrained mobileNetV2 architecture to implement transfer learning. 

Besides the target detection through imaging, mobility planning is of great importance, especially in vast outdoor areas. In this regard, in \cite{ribeiro2021unmanned} a mixed-integer problem was developed to design minimal routes for UAVs and mobile recharging stations in a post-disaster situation. A similar problem has been considered in \cite{alyassi2022auto} where ML was employed to estimate the UAV's energy consumption. In \cite{alotaibi2019}, through the collaboration of multiple UAVs, and with the goal of minimum mission time and rescuing a maximum number of people, a layered SAR algorithm was proposed. Considering fixed wing drones, \cite{keller2017coor} develops trajectory planning to conduct persistent surveillances. Also, \cite{huang2022online} considers a trajectory planning problem to obtain the for mobile objects surveillance while considering UAV's covertness.
 
\textcolor{black}{SAR operations using multiple UAVs have been also considered in \cite{zhu2020an,zhu2021a,rodriguez2020wild,de2021a}. The authors of \cite{zhu2020an,zhu2021a} use a decentralized partially observable Markov decision processes (Dec-POMDP) framework to model the uncertainty of target's location, with and without having GPS signals, respectively. \cite{rodriguez2020wild}
considers a multi-robot task allocation problem for SAR missions when both UAV and unmanned ground vehicles (UGVs) are utilized. Task allocation among multiple UAVs in SAR applications was considered in \cite{de2021a} too. 
}

In wildfire management, a coordination problem was considered in \cite{phan2008} where UAVs were used along with unmanned ground vehicles (UGVs) to fight the fire front.
Furthermore, investigating optimal UAV coalition to fully cover the area, spectrum sharing plus cell assignment, optimal number of UAV and IoT devices for a maximum detection probability, are of the other problems that have been considered in wildfire detections \cite{afghah2019wildfire,alireza2020an,osama2021the}.

\textcolor{black}{Unlike the existing literature in most of which either object detection or path planning was investigated, in this paper, a unified framework is provided including both aspects of the problem.} In particular, in this paper, the goal is to design a path planning algorithm that given the probability distribution function (PDF) of the target's location, it detects and finds the object while minimizes the average time duration by prioritizing the areas with the highest probability. \textcolor{black}{It is also worth noting that no specific model is assumed for the UAVs mobility as considered in \cite{zhu2020an,zhu2021a}. Hence, the proposed algorithm in this paper can be considered a general setup. Furthermore, as far as fire detection is concerned, a fire spot in its initial states is considered where the fire has not propagated widely yet. Therefore, new datasets were provided to train the detection algorithm. Finally, in addition to the theoretical analysis, experimental implementation has been conducted for performance evaluation. Both the analysis and the experimental results show that the proposed algorithm outperforms previously investigated mobility algorithms.}


\section{Scenario and System Model}\label{system_model}

\subsection{Simplified Settings and Intuition}
In this part, the intuition behind the problem considered in this paper is elaborated. In fact, through a simplified scenario and a Naive algorithm, it is shown that there is a need to provide a general algorithm that is capable to efficiently find the target when the location is stochastically known.

To this end, a scenario is considered where a target (e.g., a lost hiker or a fire spot) needs to be found within an area $S \subset \mathbb{R}^2$ partitioned into roughly equal cells. The location of the target known probabilistically is somewhere among the cells, i.e., its PDF is known or can be estimated through information sources. If no information is available, a uniform PDF would be assumed. A UAV is employed to fly over the region and search for the target based on the given PDF. The UAV uses its camera to monitor the area and detect the target. It is crucial that the target is found within the shortest amount of time, as any delay in finding the target could have irrecoverable and disastrous consequences. Thus, the most important goal is to minimize the \emph{detection time}.

Intuitively, to find the target as sooner as possible, one would be interested in visiting the cells with the higher probability sooner. However, this implementation is challenging as the search path must be a continuous path suitable for UAV's flying. Besides, the path must be chosen to be efficient in terms of energy consumption. Hence, an important question is that whether visiting the cells simply based on the their probabilities from the highest to the lowest one, leads to an efficient path planning algorithm in terms of detection time or not.
Through the following examples, a few insights are provided on the above question.

\begin{itemize}
	\item	\textbf{Simplified Scenario:}
\end{itemize}

Consider a drone is supposed to find a target in an area partitioned into cells denoted by index $i$ where $i = 1, 2, ..., M$. The missed detection probability denoted by $e_d$ is defined as the probability that the drone has reported no target while the target is indeed in the cell. Initially, no false alarm is considered, i.e., the false alarm probability denoted by $e_f$ is $0$. In this simplified scenario, the time required to be passed when the drone is traveling from its current cell to another one is not considered. In other words, the time amount passed from one cell to any other cell is exactly one unit of time regardless of the distance between the two cells. 

 The random variable that the target is in cell $i$ is denoted by $I$. In other words, $p_i = P(I = i)$ denotes the probability that the target exists in cell $i$. Furthermore, without loss of generality, it is assumed that
$p_1 \geq p_2 \geq, \dots, \geq p_M$. The assumptions for the simplified scenario are summarized as below:
\begin{itemize}
	\item One UAV is deployed for the operation.
	\item The probabilities of cells are of the form $p_1 \geq p_2 \geq, \dots, \geq p_M$.
	\item The UAV simply visits the cells considering the probabilities order. 
	\item No probabilities update is considered.
\end{itemize}

With this scenario and the idea that intuitively the most probable cells are selected first to search in,  the UAV is assumed to choose the cells with respect to their probabilities. With this assumption, the UAV visits $i$-th cell  at times $t_i = Mk+i$, where $k = 0, 1, \dots, $ and $i = 1, 2, \dots, M$. This is because the UAV will start over again to look for the target in the same order if it has not detected anything in the previous round and so on. If the amount of time spent by the drone to find the target is represented by $T$, with this scenario in hand, the average time is obtained in the following lemma.

\begin{lem}\label{lem:simplified}
	The average time of the simplified scenario is obtained as 
	\begin{align}
		E[T] = M \left(\frac{1}{1 - e_d} - 1\right) + E[I],
	\end{align}
\end{lem} 
where $E[.]$ denotes the expectation.
\begin{proof}
	For the proof, first note that given $I = i$, the random variable $T$ can be written as 
	\[T\mid (I = i) = M (Y - 1) + i,\]
	where $Y \sim Geometric(1 - e_d)$. Now from linearity of expectation, it is concluded that 
	\[E[T|I = i] = M\left(\frac{1}{1 - e_d} - 1\right) + i,\]
	and from the law of total expectation, $E[T]$ can be obtained as below
	\[E[T] = E[E[T|I = i]] = M\left(\frac{1}{1 - e_d} - 1\right) + E[I].\]
\end{proof}
Using Lemma \ref{lem:simplified}, $E[T]$ can be obtained for the worst case scenario where $p_1 = p_2, \dots, p_M = \frac{1}{M}$. In fact, for the worst case scenario, $E[T]$ is as follows
\begin{align*}
	E[T] & = M\left(\frac{1}{1 - e_d} - 1\right) + \frac{M+1}{2}\\
	& = \frac{M(1+e_d)}{2(1 - e_d)} + \frac{1}{2}. \numberthis \label{eq:upperbound}
\end{align*}
Equation \eqref{eq:upperbound} is actually an upperbound for $E[T]$, since given any distribution for $I$, $E[I]$ is bounded as below
\[1 \leq E[I] \leq \frac{M+1}{2}.\]

\textcolor{black}{
If $e_f \neq 0$, whenever a false alarm happens, it can be assumed that the cell is searched by a backup team and since it was a false alarm and nothing is found, an excess delay of $\Delta_f$ is added to the operation's time. Hence, given $I = i$ and $Y = k$, the false alarm delay denoted by $Z_k$ is a Binomial random variable i.e.,  $Z_k \sim Binomial((k-1)(M-1) + i - 1, e_f)$. Therefore, 
$T = Z_k + (k - 1) M$ and,}
\textcolor{black}{
\begin{align*}
P(Y = k | I = i) = P(T = Z_k + (k - 1)M + i| Y = k, I = i).
\end{align*}}
\textcolor{black}{Finally, since $Y$ and $I$ are independent, the expectation of $T$ is obtained as}
\textcolor{black}{\begin{align*}
E[T] &= E\left[ E[T|Y, I]\right] \\
& = \left((E[Y] - 1)(M - 1) + E[I] - 1\right)\Delta_f e_f \\ & + (E[Y] - 1)M + E[I] \\
& = \left(\left(\frac{1}{1 - e_d} - 1\right)(M - 1) + E[I] - 1\right)\Delta_f e_f \\
& + \left(\frac{1}{1 - e_d} - 1\right)M + E[I],
\end{align*}}
\textcolor{black}{in which the first term in the last equality represents the delay added by the false alarm.} 

The simplified scenario has two main problems: First, it is not optimal in terms of $E[T]$ since at each time that it visits a cell and nothing is detected, it chooses its next destination simply considering the same probability order. Whereas when the UAV does not detect the target in the cell with the highest probability, the probabilities of cells need to be updated according to the $e_d$. The second problem is that it does not consider a continuous trajectory. In fact, it virtually assumes that the UAV simply jumps from its current cell to the cell with the next highest probability which is not feasible in practice.

In this paper, these problems are addressed by proposing a path planning method based on updating the probabilities after each observation. 

Before going to the next section, another problem is presented which occurs when the probabilities are not updated but the amount of time required to detect the algorithm matters. 

\begin{itemize}
	\item  \textbf{Naive Algorithm:}
\end{itemize}

\begin{figure}[t]
	\centering
	\begin{tabular}{@{}c@{}}
		\includegraphics[width=.8\linewidth]{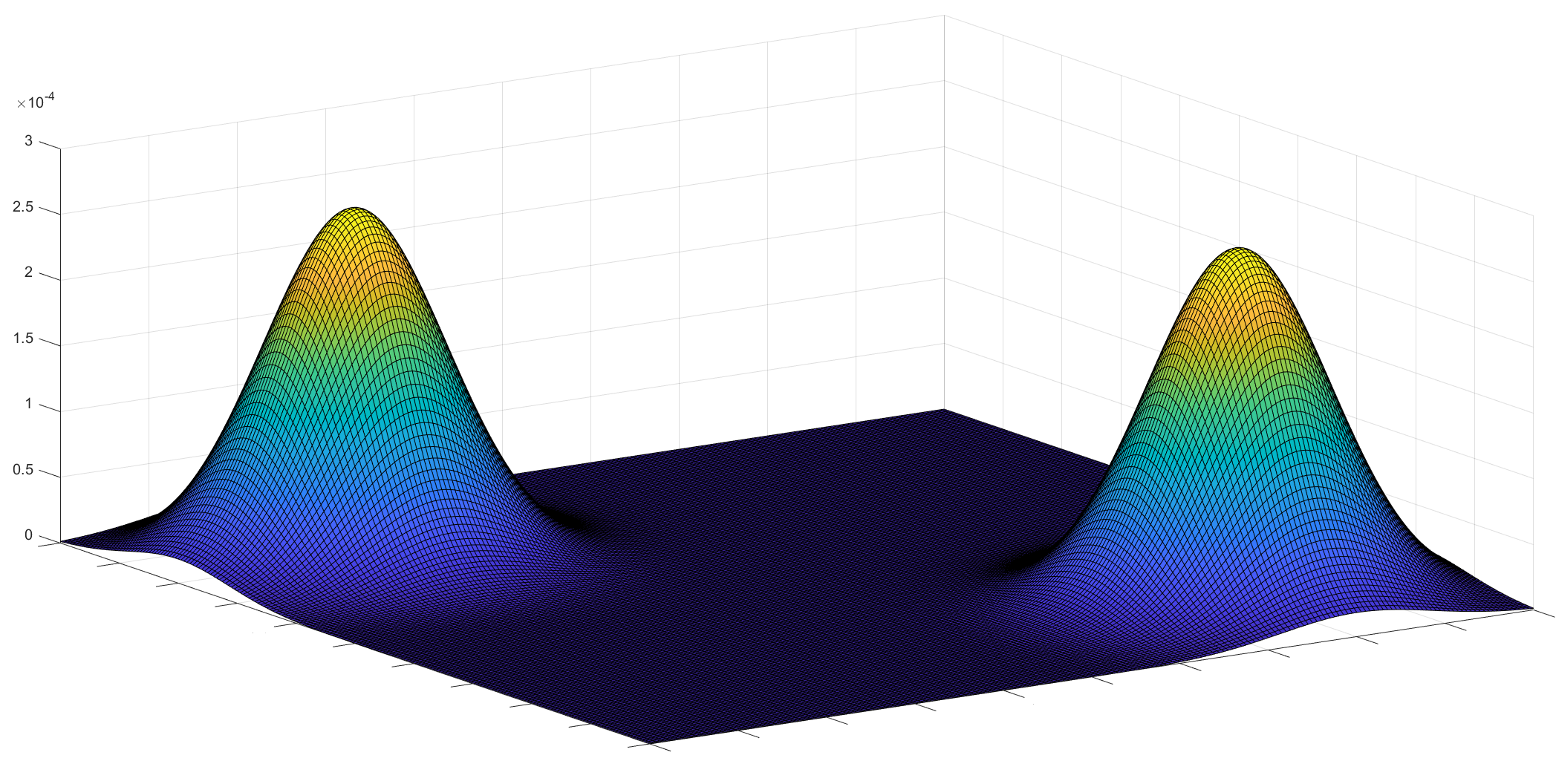} \\[\abovecaptionskip]
		\small (a) Probability distribution of the target's location.
	\end{tabular}	
	\vspace{\floatsep}	
	\begin{tabular}{@{}c@{}}
		\includegraphics[width=.8\linewidth]{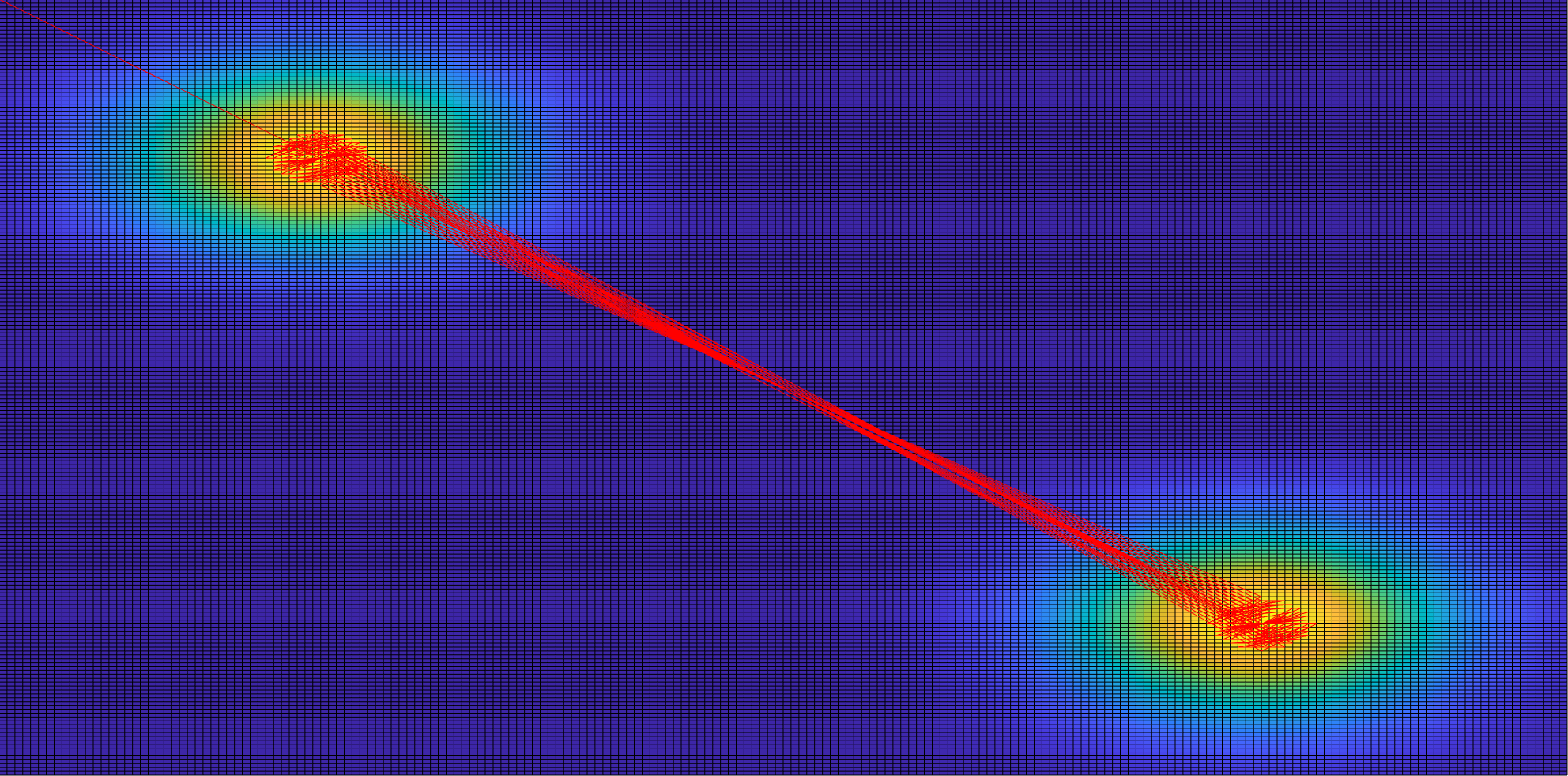} \\[\abovecaptionskip]
		\small (b) UAV's path for first 300 visits.
	\end{tabular}	
	\caption{UAV's path based on Naive algorithm for a bimodal Gaussian PDF. }\label{fig:naivealg}
\end{figure}

Now assume that the PDF is a bimodal Gaussian distribution shown in \ref{fig:naivealg}(a). In the Naive algorithm, the drone visits the cells based on the cells' probabilities from the highest to the lowest one without probability update. The assumptions for this scenario are listed as below:

\begin{itemize}
	\item One UAV is deployed for the operation.
	\item The probabilities of cells follow a bimodal Gaussian distribution
	\item The UAV simply visits the cells considering the probabilities order. 
	\item No probabilities update is considered.
\end{itemize}

 The first 300 visits are shown in  \ref{fig:naivealg}(b). As can be seen, the drone spends most of its flight time between the two peaks and misses to detect other regions resulting in a highly inefficient path planning algorithm. 


It can be seen that neither the simplified scenario nor the Naive algorithm, although intuitive, can provide efficient performance.

 In the next sections, the system model is presented in more detail.
\subsection{Scenario and Assumptions}
Suppose a drone is in charge of finding a missing person or a fire spot in the area.
The area is divided into a grid of cells, each of which has a probability of the target existing in it.
Note that this probability, in case of finding the missing person, can be obtained from utilizing some interpretation of missing person behavior, and in case of detecting the wildfire, can be obtained from the analysis of the area. \textcolor{black}{In particular, by considering hiking trails maps in the former and noticing the vegetation coverage and recognizing frequently visited areas in the latter, a prior information can be obtained. In this regard, in \cite{san2018intelligent} the authors estimate a risk/occupancy value for each cell of the area using different source of data. 
Therefore, considering a PDF for the object location is potentially a valid assumption which leads to a more efficient detection algorithm.}
Nevertheless, if there is no prior information available, the initial probability is assumed to follow a uniform distribution across the region.


\textcolor{black}{Now to efficiently detect a target, two collaborating algorithms are developed: The first algorithm is the object detection algorithm using ML. And the second one is the path planning algorithm, input of which is the output of the ML algorithm.}

It is assumed that the UAV visits one cell at each time step $t$. Hence it takes one time-step to go from one cell to one of its neighbors. It is also assumed that if a detection is reported at a cell, the rescue team will search that cell to find the target. Note that the delay induced by the rescue team is not considered. If there is no target, a false alarm will be reported and the probabilities are updated accordingly. The detection process will continue until the UAV's energy is depleted or the target is found. Note that during the path planning algorithm, cells may be visited multiple times since the probabilities are updated regularly.

The problem considered in this paper could be thought of as a variation of the traveling salesman problem and is known to be NP-Hard\cite{sokkappa1991cost}. To find a globally optimal solution requires considering each possible action and each possible observation, which implies that the cost of a solution grows exponentially with the number of cells and actions. Hence, sliding window technique with a length of $W$ is used to find a sub-optimal path from the current location.

The general idea of the path planning algorithm is to search the most likely cells first; however, there is a trade-off between the time to reach a cell and the success probability of visiting that cell. The idea is that visiting the cell with the maximum probability may be less attractive than visiting the adjacent cells because it takes a long time to fly to the cell which has the maximum probability, but it may slightly improve the success rate. Therefore, we design a windowing-based path planning, the details of which will be discussed in Section \ref{sec:path-planning}.

\subsection{Area Decomposition}
In the first step, the area should be segmented through an approximate cellular decomposition. In this method, the area is divided into rectangles, and a point is placed in the center of each rectangle. The UAV is assumed to fully cover the cell when it reaches that particular point. The size of the rectangles is calculated from the field of view (FOV) of the camera.

Camera's FOV is the area covered by UAV’s camera when it is flying at altitude $H$. The size $(w,l)$ of the projected area can be obtained by the following equations
$$w = 2H.\tan\left(\frac{\alpha}{2}\right),$$
$$l = 2H.\tan\left(\frac{\beta}{2}\right),$$
where $\alpha$ and $\beta$ are the vertical and horizontal angles of the camera, respectively.

\begin{figure}[htb]	
	\centering
	\includegraphics[width=0.5\textwidth]{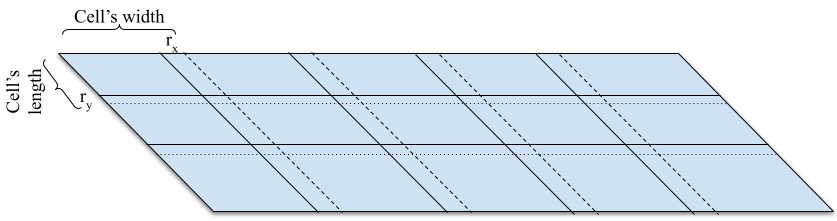}
	\caption{Projected areas with overlaps. Centers of the rectangles are the path waypoints.}
	\label{fig.projected_area}
\end{figure}

Therefore, the area of interest is decomposed into a sequence of rectangles, which is denoted by $C_i$ where $1 \le i \le M$. Then, the drone must be programmed to fly over each subregions' center. The complete path is then stored as a list of coordinates, called waypoints, and the drone moves from one waypoint to the next until it finds the missing object or it is out of energy. To cover the area completely, projected areas must overlap as shown in Figure \ref{fig.projected_area}. The amount of overlap can be chosen and can vary on each side. The horizontal and vertical overlaps are denoted as $r_x$ and $r_y$, respectively.

At every UAV flight time step, the camera footprint of the search area is treated as a glimpse. This way, the search area can be discretized, and the UAV path-planning problem can be modeled as a discrete combinatorial optimization problem.


\subsection{System Model}
Since there is uncertainty and randomness in the operation (i.e., the location of the target is not predetermined, and could follow a known or unknown statistical distribution), it is essential to consider a probabilistic view. More specifically, let the underlying probability space be represented as $(\Omega, \mathcal{F}, P)$, where $\Omega$, $\mathcal{F}$, and $P$ represent the sample space, the event space, and the probability function, respectively. 

The main assumptions for the system model are as below:

\begin{itemize}
	\item One UAV is deployed for the operation.
	\item The probabilities of cells follow an arbitrary distribution model.
	\item The UAV visits the cells based on probability orders while considering the time constraint.  
	\item Probability update is considered.
\end{itemize}


Now the first step is to determine the probabilistic map which combines all available information including past experiences. Each cell $i$ where $i = 1, 2, \dots, M$ in the probabilistic map has a probability of $P(I_i)$ which shows the probability that the target exists in that cell. Thus, with this definition, it is concluded that
\[\sum_{i=1}^M P(I_i)=1. \]




It is assumed that first the drone makes a new observation at time step $t$ and then the ML algorithm determines whether the target is in the captured image or not. In this regard, $D_i$ denotes the target detection random variable by the ML algorithm upon visiting the cell $i$. 


Note that even if the drone captures an image of the target, it is possible that the object will not be detected, such possibilities are referred to as \textit{missed detection probability} which is denoted by $e_d$ and is defined as
\begin{equation}\label{eq:missDet}
\begin{aligned}
P(D_i^c | I_i)= e_d,
\end{aligned}
\end{equation}
where $I_i$ shows the existence of target in $i$-th cell. Missed detection may be due to the inefficiency of the detection algorithm or a defect in the camera. 

On the other hand, it is possible that the detection algorithm misidentifies the object in the image captured by the UAV. These possibilities are referred to as \textit{false alarm probability} which is denoted by $e_f$ and defined as
\begin{equation}\label{eq:falsAla}
	\begin{aligned}
		P(D_i | I_i^c)= e_f,
	\end{aligned}
\end{equation}
where $I_i^c$ shows the nonattendance of target in the $i$-th cell.

In the next section, the probability updating framework and the path planning algorithm are provided.

\section{Probability Updating and Path Planning Algorithm}
\label{sec:algorithm_SAR}

\subsection{Updating Target Detection Probabilities}
\label{sec:update}


To update the detection probabilities for the cells, the Bayesian rule is utilized using the information obtained from observations. 
At each time step $t$, the conditional updated probability at cell $i$ is obtained as below:
\begin{align*}
	P(I_i | D_i^c) &=\frac{P(D_i^c|I_i)P(I_i)}{P(D_i^c)} \\
	&=\frac{P(D_i^c|I_i)P(I_i)}{P(D_i^c|I_i)P(I_i)+P(D_i^c|I_i^c)P(I_i^c)}. \numberthis \label{eq:errorpdetection}
\end{align*}
Now, using Equations \eqref{eq:missDet} and \eqref{eq:falsAla}, one can obtain
\begin{align}
	P(I_i | D_i^c)& = \frac{e_dp_i}{e_dp_i+(1-e_f)(1-p_i)}\\
	& = \frac{e_dp_i}{b_i}, \numberthis \label{eq:errorpdetection2}
\end{align}
where $p_i\stackrel{\triangle}{=}P(I_i)$ and $b_i \stackrel{\triangle}{=} e_dp_i + (1-e_f)(1-p_i)$. 
Similarly, the probability map is updated for other cells, i.e.,  $1 \le j \le M$ where $j \neq i$, as below
\begin{align*}
	P(I_j | D_i^c)&= \frac{P(D_i^c|I_j)P(I_j)}{P(D_i^c)}\\
	&= \frac{(1-e_f)p_j}{b_i}.\numberthis \label{eq:Condi-updat-other-cell}
\end{align*}
Therefore, after visiting the $i$-th cell and conditioned on $D_i^c$, the probability of each cell can be updated as below:
\begin{equation}\label{eq:updateprob}
	\begin{aligned}
		P(I_j)=\begin{cases}
			p_i\frac{e_d}{b_i} & j=i\\
			p_j\frac{1-e_f}{b_i} & j \neq i 
		\end{cases}     .
	\end{aligned}
\end{equation} 


\subsection{Windowing-based Path Planning Algorithm}
\label{sec:path-planning}

Now the details of the proposed path planning algorithm is provided in the sequel. The goal of this algorithm is to minimize the average time of finding the target, ${E}[T]$. Assume that $B_t$ denotes the target detection at time step $t$. Therefore, $f_t$ defined as the probability that the target is detected for the first time at time step $t$, can be obtained as follows
\begin{equation}
	\begin{aligned}
		f_t = \left(1-P\left(B_t^c|B_{1:t-1}^c\right)\right)\prod_{j=1}^{t-1}P\left(B_j^c|B_{1:j-1}^c\right).
	\end{aligned}
\end{equation}
Assuming no false detection reported from the ML algorithm, the expectation of the number of steps required to detect the target can be obtained as  
\begin{equation}\label{eq:errorpdetection3}
	\begin{aligned}
		{E}[T] = \sum_{t=1}^{\infty} t.f_t,
	\end{aligned}
\end{equation}
where $t \in N$ is the number of time steps to detect the target for the first time.

Note that this problem is a variation of the traveling salesman problem and is known to be NP-Hard\cite{sokkappa1991cost}. To find a globally optimal solution requires considering all the possible actions and observations, which implies that the cost of a solution grows exponentially with the number of cells and actions. Rather than finding a globally optimal path, the sliding window technique with length $W$ is used to find the sub-optimal path from the current location. It should be noted that in order not to compromise the global optimality of trajectory, after visiting each cell and updating the probability map, the proposed algorithm again finds a sub-optimal path with length $W$ from the current location. In other words, it is desirable to visit each cell of the window in a way that minimizes the expected time of the target detection by minimizing $E[T]$.

Note that if the target exists in the cells with almost an equal probability, it is better to visit the cells in a zigzag manner instead of prioritizing to visit the cell with the maximum probability. Because in this case, it is easy to realize that the proposed algorithm would take a longer time while only improves the success probability slightly. 

\begin{figure}[htb]	
	\centering
	\includegraphics[width=0.4\textwidth]{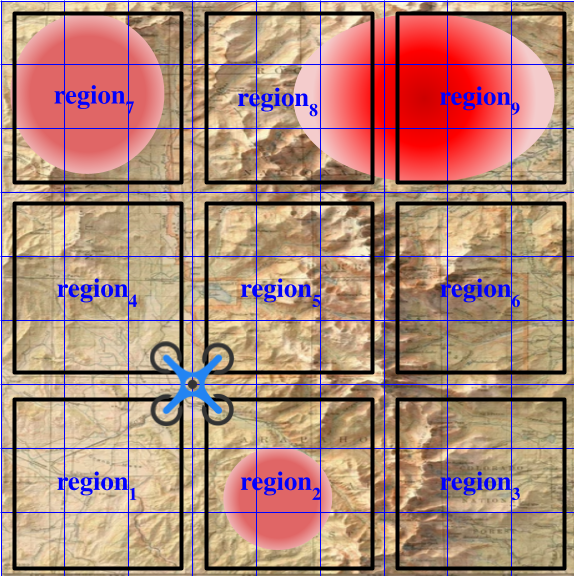}
	\caption{Projected areas with overlaps. Centers of rectangles are the path waypoints.}
	\label{fig.region}
\end{figure}

In order to make a tradeoff between the time to reach a cell and the success probability of visiting that cell, $W$ by $W$ cells are aggregated to form non-overlap regions. Therefore, denoting by $P(R_l)$ the probability of the target existing in the region $l$, $P(R_l)$ is obtained as
\begin{equation}\label{eq:defineregion}
	\begin{aligned}
		P(R_l) = \sum_{k=1}^{W^2} p_{((l-1)W^2+k)},
	\end{aligned}
\end{equation}
where $1 \le l \le \frac{M}{W^2}$. 
Now the distance between the regions $R_l$ and $R_k$ denoted by $d_{R_l,R_k}$ is obtained from the distance between the centers of these two regions. Figure \ref{fig.region} shows an area of interest partitioned to $9$ regions where each region includes $3\times3 = 9$ cells, i.e., $W=3$. 

Let $R_{\text{max}}$ denote the region with the highest probability among all the regions. In other words,
\begin{equation}\label{eq:argmaxregion}
	\begin{aligned}
		R_{\text{max}} = \arg \max_{l \in [1:\frac{M}{W^2}]} {P(R_l)} . 
	\end{aligned}
\end{equation}
Furthermore, let $R_{\text{max}}^{\text{local}}$ denote the region with the highest probability among the adjacent regions of the current one written as below
\begin{equation}\label{eq:argmaxregion1}
	\begin{aligned}
		R_{\text{max}}^{\text{local}} = \arg \max_{l \in A(R_c)} {P(R_l)} , 
	\end{aligned}
\end{equation}
where $R_c$ denotes the current region and $A(R_c)$ is the set of adjacent regions for the current region. Finally, let $R_p, p \in A(R_c)$ indicate the region located in a straight line between $R_c$ and $R_{max}$. Now the next region, denoted by $R_{n}$ is chosen as below:
\begin{align}
	R_{n} = 
	\begin{cases}
		R_{\text{p}} &
		\frac{P\left(R_{\text{max}} \right)}{P\left(R_{\text{max}}^{\text{local}}\right)} > \frac{d_{R_c,R_{\text{max}}}}{d_{R_c,R_{\text{max}}^{\text{local}}}}\\
		R_{\text{max}}^{\text{local}} & \text{Otherwise}
	\end{cases}.
\end{align}

Note that by the next region, it is meant the next region chosen as the next destination that the UAV is supposed to fly to. It is obvious that in order to fly to the next region, the UAV first may visit some middle cells in the subsequent time steps. 

Intuitively, with this comparison, it can be determined whether flying to the region with the maximum probability is attractive given the time it takes to get there or not. Therefore, in this step, it is decided which region should be visited next. This step will be repeated after $W$ observations.

After selecting the region, cells' order is investigated on the selected region. If the cell index chosen to be visited at time $t$ is denoted by $o_t$, the next cell index is denoted by $o_{t+1}$ which is an adjacent cell of the current cell, $o_t$, and can be selected as below
\begin{equation}\label{eq:argminimizeT}
	\begin{aligned}
		o_{t+1} = \arg \min_{a \in A(o_t)} {E[T_a]} , 
	\end{aligned}
\end{equation}
where $A(o_t)$ is the set of adjacent cells for the current cell. Furthermore, $E[T_a]$ is obtained from the following equation
\begin{equation}\label{eq:minimizeT}
	\begin{aligned}
		E[T_a]& = E\left[T|D_{a:a+W}\right]+E\left[T|D_{a:a+W}^c\right]\\
		&= \sum_{i=1}^W t_ip_{a+i} +  (M-W)\left(1-\sum_{k=1}^Wp_{a+k}\right),
	\end{aligned}
\end{equation}
where $M$ is the maximum time to visit all the cells in order. Therefore, one of the adjacent cells of the current cell which minimizes Equation \eqref{eq:minimizeT} is chosen. Intuitively, it means that the cells with higher probabilities are visited sooner\footnote{\textcolor{black}{It is worth mentioning that there might be obstacles in the path to the next cell determined by Algorithm \ref{alg:pathplanningalgorithm}. In this situation, the collision avoidance has been taken care of in the implementation phase where the UAV uses its sensors to avoid accidents with obstacles.}}. Algorithm \ref{alg:pathplanningalgorithm} represents the proposed path planning method for the target detection.

\begin{algorithm}
	\begin{algorithmic}[1]
		\footnotesize	
		\Function{PathPlanning}{$P, w, O, m$}	
		\Inputs{
			$P$ probabilistic map\\
			$w$ is window size\\
			$O$ start position\\
			$m$ is the location of the target}
		\Output{Time to find target}
		\State Construct non-overlapping regions by each $w\times w$ cells; call these regions: $\text{R}_1$, $\text{R}_2, ..., \text{R}_Z$, $Z = \frac{M}{w^2}$
		
		
		\State $current.region = O$
		\While{no target is found}
		\For{$w$ observations}
		\State Among all straight paths from $o_{t}$, select $o_{t+1}$ that minimizes Eq. \eqref{eq:minimizeT}
		\State $observation$ = algorithm \ref{alg:DroneMLrecognition}($o_{t}, o_{t+1}$)
		\If{$observation$ $!= 0$}
		\State Report to ground control station 
		\EndIf		  
		\State Update probability map using Eq. \eqref{eq:updateprob}
		\EndFor
		\State Find the highest probability of all regions $R_{\text{max}}$
		\If{$\frac{P\left(R_{\text{max}} \right)}{P\left(R_{\text{max}}^{\text{local}}\right)} > \frac{d_{R_c,R_{\text{max}}}}{d_{R_c,R_{\text{max}}^{\text{local}}}}$}
		\State $R_{n}$ = $R_{\text{p}}$
		\Else
		\State $R_{n}$ = $R_{\text{max}}^{\text{local}}$		
		\EndIf
		\EndWhile		
		\EndFunction
	\end{algorithmic}
	\caption{Path Planning}
	\label{alg:pathplanningalgorithm}	
\end{algorithm}

So far, only the path planning algorithm has been developed for which the inputs of the making decisions were originated from the ML algorithm. Therefore, the ML algorithm is developed in the next section.

\subsection{Machine Learning Architecture}

\begin{figure*}[h]
	\centering
	\includegraphics[width=\textwidth,height=3cm]{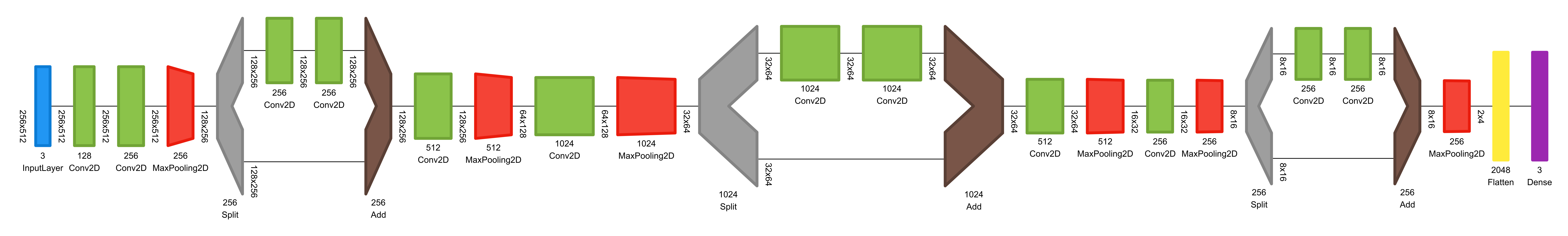}
	\includegraphics[width=0.6\textwidth]{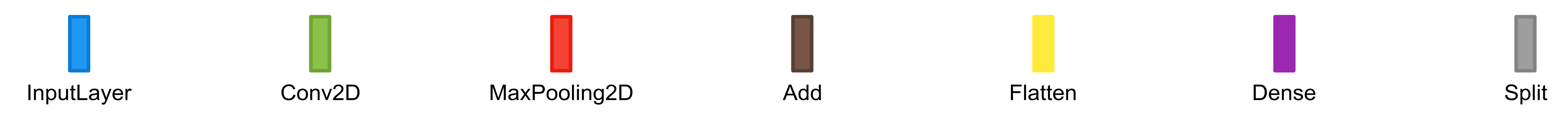}
	\caption{The architecture of the ResNet. Visualization has been created using \cite{net2vis}.}
	\label{fig.MLModel}	
\end{figure*}

Generally to solve intelligent decision-making problems, deep neural networks  are used since they not only can learn patterns from the raw data themselves but also are able to learn the features automatically by adding more layers in order to learn more complex features. However, they suffer from the vanishing gradient problem \cite{hochreiter1998vanishing}, where the gradients become vanishingly small when training using gradient-based methods and backpropagation. \textcolor{black}{In this regard, Residual Neural Network (RestNet) introduced by He et al \cite{he2015resnet}, is a class of deep neural networks built for dealing with the vanishing gradient problem, especially for image classification. ResNets approach this problem by adding shortcut connections that skip one or more layers, such that the input of the skipped layers is added to the output of the skipped layers. As a result, it has obtained a  3.57\%  error  on  the  ImageNet  test set as well as won first place in the ImageNetdetection, ImageNet localization, COCO detection, and COCO segmentation in the 2015 ILSVRC and COCO competitions  \cite{ILSVRC15}, \cite{coco2014data}. Therefore, this paper utilizes a ResNet architecture, details of which are provided in the sequel.}

\textbf{Data: } Experiments are conducted for fire and missing person detection in snow. In case of fire detection, the dataset consists of 2646 images divided into 3 classes: \emph{Fire}, \emph{Smoke}, and \emph{Spare}. Spare is considered any other object that the UAV might encounter such as trees, houses and the ground. Out of 2646, 987 are the Fire, 781 are the Smoke, and 818 are Spare. In case of the lost hiker detection, the dataset consists of 1400 images divided into 2 classes: a \emph{Snow-Covered Victim}, and \emph{Spare} (any unwanted object such as trees, houses, and snow-covered ground). \textcolor{black}{Out of these 1400 images, 650 are the Snow-Covered Victim and the rest are Spare.} \textcolor{black}{In both datasets, roughly 15\% of the images are taken by the drone at heights between 15-30 meters during some preliminary flights to help the model generalize better to the real world. The rest of the images are carefully and manually selected from online repositories and available datasets \cite{aiformankind} \cite{dincer2021data} in a way that they best represent different classes of the data in the real world setting as well.}

\textcolor{black}{Furthermore, for the validation procedure, an addition of 300 images is taken by the drone at a second location. Finally, testing is performed at two other new locations. In other words, different locations are used for the three datasets – training, validation, and testing – to ensure that the experiment is authentic and that the model is truly generalizing rather than overfitting to the training or validation data.}


\textbf{Model: } As discussed earlier, the model used in this paper is a ResNet, the details of which are highlighted in Figure \ref{fig.MLModel}. The designed ResNet has 6 convolution and 3 residual blocks, or equivalently 12 convolutional layers, 12 activation layers, and 6 pooling layers. Each convolutional block consists of a sequential and ordered arrangement of a convolutional layer, a batch normalization layer, rectified linearity activation (ReLU) layer, and a maximum pooling layer. A residual block, on the other hand, consists of two sequential convolutional blocks (minus pooling) that are added to the input of the first convolutional blocks.

\begin{figure}[htbp]	
	\centering
	\includegraphics[width=0.45\textwidth]{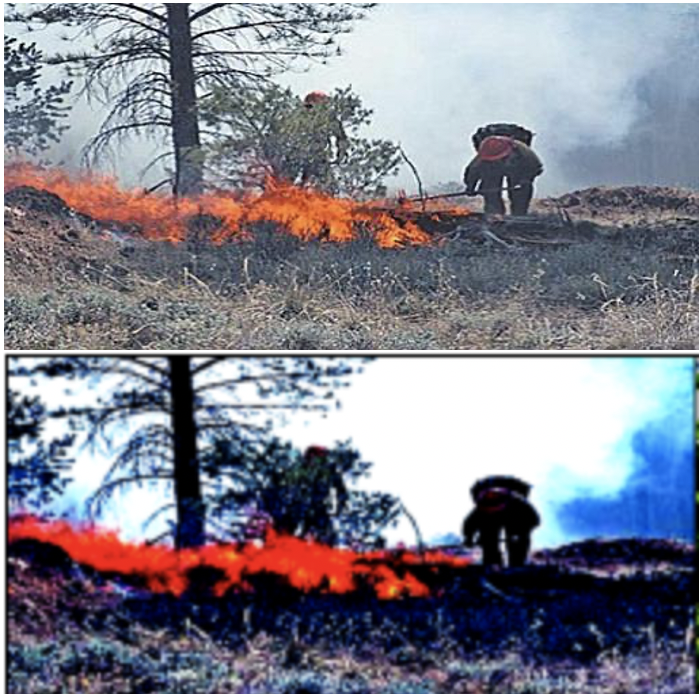}
	\caption{Data Normalization is applied on the input images to highlight the  key features in the images.}
	\label{fig.firedatanorm}
\end{figure}

\begin{figure}[htbp]	
	\centering
	\includegraphics[width=0.45\textwidth]{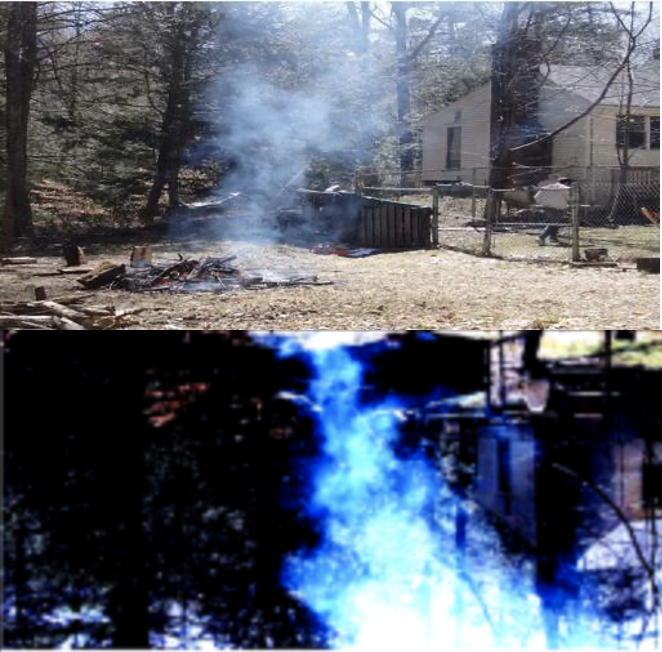}
	\caption{In addition to Data Normalization, Data Augmentation is applied to the training data in order to increase new images in each epoch.}
	\label{fig.smokedataaug}
\end{figure}

\vspace{0.5 cm}
\textbf{Training \& Validation: } The model is trained for 8 epochs using an Adam Optimizer and a learning rate scheduler with the Cross Entropy Loss Function. \textcolor{black}{A detailed list of training parameters has been provided in Table \ref{tab:summary}. It is worth mentioning that a lower learning rate is preferred due to the nature of the data as a larger learning rate would cause jumps around the global minima.}

\begin{table}[t!]
	\centering
	\caption{\textcolor{black}{Summary of Training Parameters.}}
	\textcolor{black}{
	\begin{tabular}{c|c|}
\hline
		\multicolumn{1}{|c|}{\textbf{Parameter}}  &  \multicolumn{1}{c|}{\textbf{Value}} \\ \hline	
		\multicolumn{1}{|c|}{Image Size} & 256 $\times$ 512 $\times$ 3 \\ \hline
		\multicolumn{1}{|c|}{Epochs} & 8 \\ \hline
		\multicolumn{1}{|c|}{Optimizer} & Adam\\ \hline
		\multicolumn{1}{|c|}{Learning Rate} & 0.0005\\ \hline	
		\multicolumn{1}{|c|}{Weight Decay} & $10^{-4}$\\ \hline	
		\multicolumn{1}{|c|}{Gradient Clipping} & 0.1\\ \hline	
		\multicolumn{1}{|c|}{Feature Extraction} & Z-normalization \\ \hline
		\multicolumn{1}{|c|}{Batch Normalization} & Yes\\ \hline
		\multicolumn{1}{|c|}{Dropout $\mathrm{p}$} & 0.3\\ \hline
		\multicolumn{1}{|c|}{Fire Validation Acc.} & 0.932\\ \hline
		\multicolumn{1}{|c|}{SAR Validation Acc.} & 0.94\\ \hline
		\multicolumn{1}{|c|}{Time to Inference (IPhone X)} & \\
		\multicolumn{1}{|c|}{including drone-device latency} & 3 s \\ \hline
	\end{tabular}}
	\label{tab:summary}
\end{table}

 \textcolor{black}{Furthermore, since a small data set is used, several data augmentation strategies such as manual feature extraction, flipping, cropping, and zooming is applied during the training process. In fact, while larger datasets are preferred for generalization, the data augmentation techniques have shown to be successful in feature extraction and synthetically increasing the size of the data.}

The training process includes the following functions:
\begin{itemize}

	\item[--] Data Normalization is used for training, validation, and the implemented model on the drone. The image tensors are normalized by subtracting the mean and dividing by the standard deviation, both of which are calculated separately. Data Normalization highlights the essential elements of the input image, making it easier for the machine learning classification. Figures \ref{fig.firedatanorm} and \ref{fig.smokedataaug} show a sample of the data normalization.

	\item[--] Data Augmentation. Note that the size of the dataset is smaller than many of the widely available datasets. Hence, in order to avoid over-fitting and ensuring that the model generalizes well to the real world, the apparent size of the dataset is increased using data augmentation. 
	To this end, the images are 1) \emph{padded on the right}, 2) \emph{randomly cropped}, and then 3) \emph{flipped with a fixed probability}. An example of data augmentation can be seen in Figure \ref{fig.smokedataaug}.

	\item[--] Batch Normalization: The data is initially normalized by subtracting the mean and dividing by the standard deviation (Data Normalization). Batch Normalization takes the same principle and applies it to each layer in the neural network in order to further extract the features in the outputs of each layer, before feeding it into the next layer \cite{batch2015normalization}.	
	
\end{itemize}
\begin{figure*}[t]	
	\centering
	\subfloat[Validation accuracy vs epoch number]{
		\includegraphics[width=0.95\columnwidth]{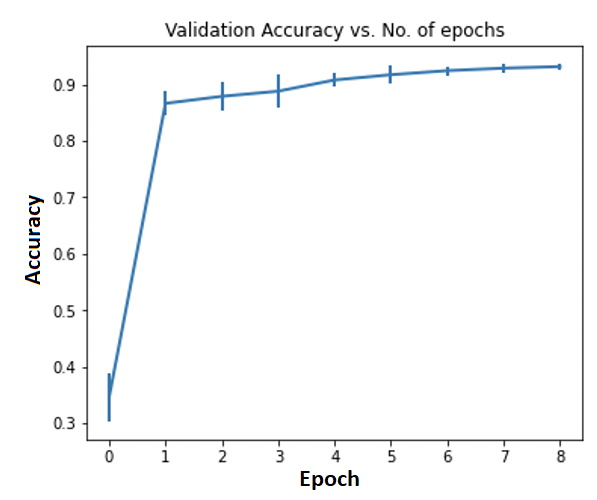}}		
	\label{fig.accuracy_curv}\hfill
	\subfloat[Training and Validation loss vs epoch number]{
		\includegraphics[width=0.95\columnwidth]{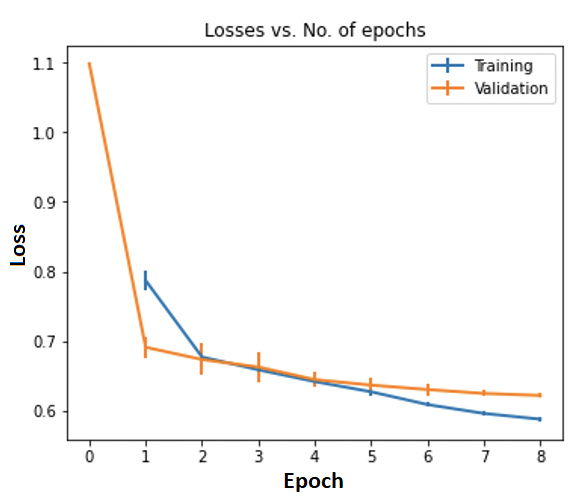}	
		\label{fig.loss_curve}}		
	\caption{Training and Validation for the model over 30 seeds. The average values are plotted while the standard deviation is used for providing confidence intervals. The plots included are: (A) Validation accuracy vs epoch number, and (B) Training and Validation loss vs epoch number.}
	\label{fig.metricvsepoch}
\end{figure*}


Besides, the machine learning model is trained for 30 seeds and the validation accuracy, validation loss, and training loss are recorded for each seed. The average validation accuracy at the end of the 30 seeds is $93.8\%$ while the average training and validation losses are 0.5857 and 0.6061, respectively. Average metrics vs epoch, with the standard deviation for confidence intervals, are represented in Figure \ref{fig.metricvsepoch}.

 \textcolor{black}{Table \ref{tab:metricMLalgo}, demonstrates the results collected on validation data. In this table, Fire has the lowest accuracy and highest precision. This is due to the size of the fire when conducting the validation tests,  where for safety, it has been kept small.}
 	
 	\begin{table}[t!]
 		\centering
 		\caption{\textcolor{black}{Preliminary Classification metrics for validation data. The low accuracy of Fire is due to the fact that the fire size has been chosen small for safety. It has been increased slightly later in Testing presented in Table \ref{tab:metricMLalgolowerheight}. }}
 		\textcolor{black}{
 		\begin{tabular}{c|c|c|c|c|}		
 			\cline{2-5}
 			&   \begin{tabular}{@{}c@{}} Precision \end{tabular} & \begin{tabular}{@{}c@{}} Recall \end{tabular} &   \begin{tabular}{@{}c@{}} F1-score \end{tabular} &  \begin{tabular}{@{}c@{}} Support \end{tabular} \\ \hline
 			\multicolumn{1}{|c|}{Fire} & 0.99 & 0.74 & 0.85 & 120 \\ \hline
 			\multicolumn{1}{|c|}{Smoke} & 0.91 & 0.95 & 0.93 & 120\\ \hline
 			\multicolumn{1}{|c|}{Spare} & 0.82 & 0.99 & 0.90 & 120\\ \hline	
 		\end{tabular}}
 		\label{tab:metricMLalgo}
 	\end{table}
Besides, it can be seen that the Fire class has the highest precision due to the fact that it has extremely few false alarms. The Smoke class has fewer false alarms as well where it has a precision of $0.91$. The Spare class, on the other hand, has the lowest precision since many misclassifications in the Fire testing are classified as Spare when the flames are not visible, giving rise to a larger number of false alarms. The highest precision for Fire and Smoke is advantageous as it is indicative of a low number of false alarms. The Fire class has a lower recall, while the Smoke and Spare classes have a significantly higher recall which implies that the Fire class has a larger number of missed detection while the Smoke and Spare classes do not. The size of the fire has been increased slightly in the experiments conducted for testing later presented in Table \ref{tab:metricMLalgolowerheight}.

\textcolor{black}{
	Finally, to find the ideal stopping point of training, an analysis of the misclassified images is conducted. It has been observed that the misclassified images consisted only in case of outliers, which particularly occurs when the different classes share features. For example, cloud and smoke, or cloud and white houses and snow share features. In this case, training beyond involves overfitting to these outliers in the training dataset which in turn leads to a decrease in accuracy on the validation data. Therefore, it is concluded that the optimal stopping point is 8 epochs.}

\vspace{0.5 cm}
\textbf{Testing:} In validation process, the lower recall for Fire class happens because it is difficult to view a campfire due to it being covered by logs and sticks. This is in contrast to viewing the flames at an angle in which the flames are more visible. \textcolor{black}{To deal with the assumption that the lower accuracy of the Fire is caused due to less visible flames, a significantly larger controlled burn has been conducted in open burning season (ending on May 1)} and it is observed that an accuracy of $93\%$ is achieved for the Fire class with the metrics as seen in Table \ref{tab:metricMLalgolowerheight}.

\begin{table}[t!]
	\centering
	\caption{\textcolor{black}{Classification metrics for testing data.}}\textcolor{black}{
	\begin{tabular}{c|c|c|c|}
		\cline{2-4}
		&   \begin{tabular}{@{}c@{}} Precision \end{tabular} & \begin{tabular}{@{}c@{}} Recall \end{tabular} &   \begin{tabular}{@{}c@{}} F1-score \end{tabular}  \\ \hline
		\multicolumn{1}{|c|}{Fire} & 0.99 & 0.93 & 0.96 \\ \hline
		\multicolumn{1}{|c|}{Smoke} & 0.97 & 0.95 & 0.96 \\ \hline
		\multicolumn{1}{|c|}{Spare} & 0.92 & 0.99 & 0.96 \\ \hline
	\end{tabular}}
	\label{tab:metricMLalgolowerheight}
\end{table}

	\textcolor{black}{In this table, due to the visible flames, it can be seen that the metrics are improved. With a higher accuracy reported with more visible flames, the prediction is that on an early forest fire, which can be essentially larger than a campfire of 1 ft tall, the algorithm even performs more desirably.}

\textcolor{black}{As mentioned earlier, to ensure authenticity of the experiment and avoid overfitting to the training or validation data, tests are conducted on new fire spots different from those of training and validation. This also allows the measurement of the generalization performance of the machine learning method, presented in Table \ref{tab:metricMLalgolowerheight}. Depending on the availability and nature of the area surrounding the fire spot, a disturbance level is added to each testing image that measures how much disturbance is being added from the surroundings. For instance, when testing for Fire or Smoke, the drone camera is adjusted to display half or full of the surrounding tree to confuse the machine learning algorithm to determine whether the image is Smoke or Spare. Furthermore, this testing has been conducted for varying amounts of environmental disturbance.}

\textcolor{black}{It is worth mentioning that the largest distance for detecting Fire has been measured to be 20.2 meters (66.3 feet) on the firepit and 28 meters (91.86 feet) on the controlled fire, while the highest distance for detecting Smoke has been 37 meters (121.39 feet).}

\textcolor{black}{Table \ref{tab:searchandrescue} presents the test results for the SAR operation considering two cases: with snowflakes and without snowflakes. In the former, falling snow is superimposed into the images in order to add noise and further test the model's limits. Also, a comparison of ResNet and CNN has been provided in this table. In this setting, in order to add diversity to the data and further test the model, participants with five different clothes colors were involved. The testing also involved the participants being in different locations in the image to even being partially present where the model has shown to be able to detect. Furthermore,
 varying amounts of snow on the ground has been considered for the model testing, ranging from being completely covered in snow to partial coverage. Finally, the largest height of detecting a person has been measured to be 24 meters (78.74 feet) with snowflakes and 27 meters (88.58 feet) without snowflakes}. 

\begin{table}[t!]
	\centering
	\caption{\textcolor{black}{Accuracy of testing data for SAR in snow.}}\textcolor{black}{
	\begin{tabular}{c|c|c|}		
		\cline{2-3}
		&   \begin{tabular}{@{}c@{}} With Snowflakes \end{tabular} & \begin{tabular}{@{}c@{}} Without Snowflakes \end{tabular}  \\ \hline
		\multicolumn{1}{|c|}{CNN} & 0.582 &  0.731 \\ \hline
		\multicolumn{1}{|c|}{ResNet} & 0.897 & 0.943 \\ \hline
	\end{tabular}}
	\label{tab:searchandrescue}
\end{table}

The entire machine learning algorithm is shown in more detail in Algorithm \ref{alg:DroneMLrecognition}. First, the UAV automatically flies from the current location to the next location. Then, it should take an image by setting its camera to shoot mode and resize the image and fetch the captured image. Finally, the ML recognition runs on the captured image and returns the detection class to Algorithm \ref{alg:pathplanningalgorithm}. 

\begin{algorithm}
	\begin{algorithmic}[1]
		\footnotesize	
		\Function{Drone and ML Action}{$c, n$}	
		\Inputs{
			$c$ := current location of the UAV\\
			$n$ := next location of the UAV}
		\\
		\Output{Target detection}		
		\State Move UAV from location $c$ to location $n$
		\If{Fetch camera}
		\State $image$ = captured image from UAV
		\State Run ML recognition($image$)
		\State return $0, 1,$ or $2$ for $Spare, Fire,$ or $Smoke$
		\Else
		\State return Fetch camera error
		\EndIf		
		\EndFunction
	\end{algorithmic}
	\caption{Drone and ML Action}
	\label{alg:DroneMLrecognition}	
\end{algorithm}

\section{Simulation and Experimental Results}

\subsection{Camera Model}
\begin{figure}[htb]	
	\centering
	\includegraphics[width=0.49\textwidth]{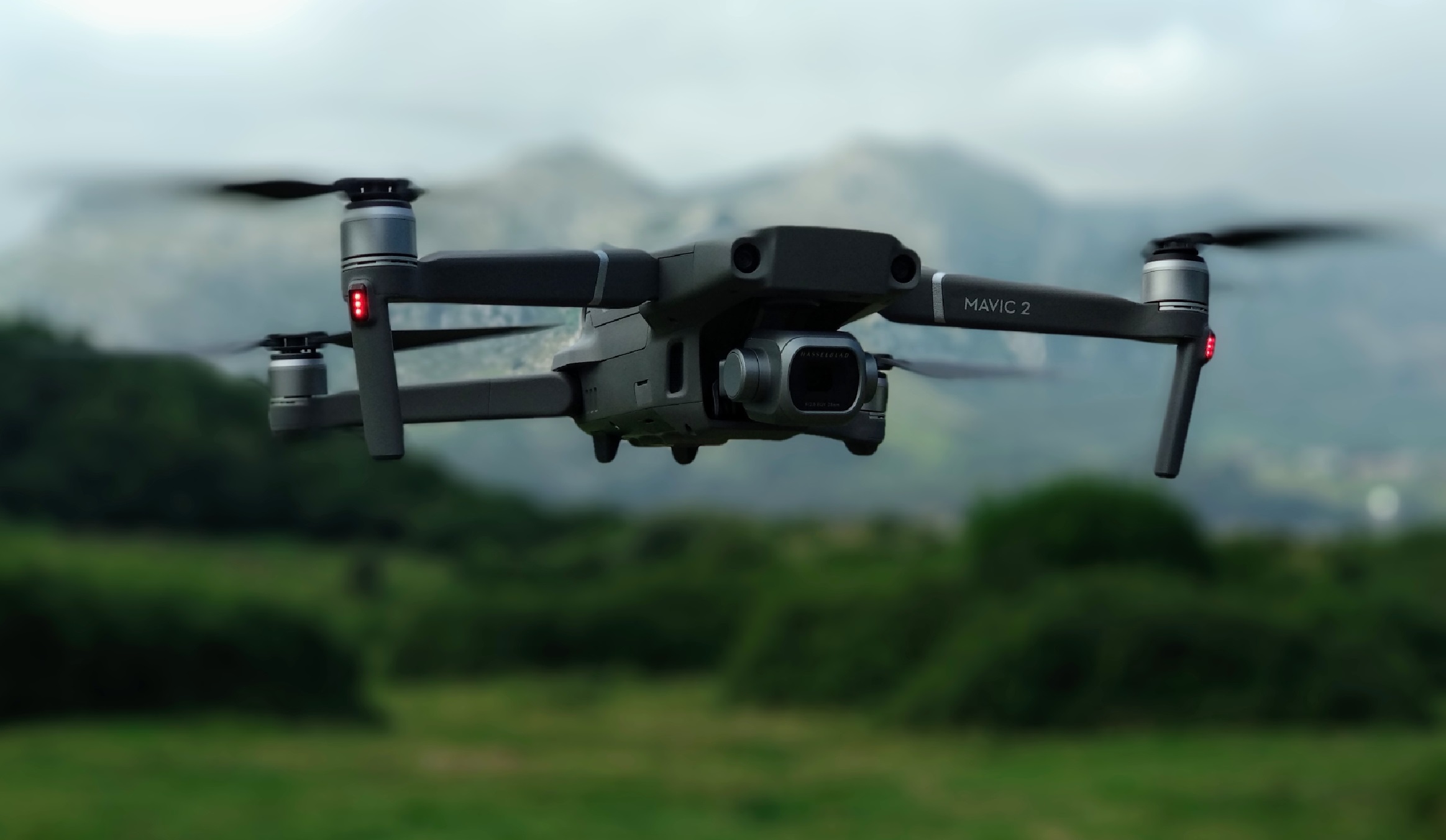}
	\caption{DJI Mavic $2$ Zoom with a 4k camera.}
	\label{fig.drone}
\end{figure}

In order to obtain the numerical results through simulation and experiments, a DJI Mavic $2$ Zoom with a 4k camera shown in Figure \ref{fig.drone}, has been used. The goal is to scan a given area and reconstruct its map with a spatial resolution $R$ greater than or equal to $R_d$, expressed in pixels/cm.

\subsection{Power consumption}
One critical issue of UAV operation is the limited onboard energy of UAVs, which renders energy-efficient UAV-based operations particularly important. The UAV energy consumption is in general composed of two main components, namely the communication related energy and the propulsion energy. Depending on the size and payload of UAVs, the propulsion power consumption may be much more significant than communication-related power. To this end, proper modeling for UAV propulsion energy consumption is crucial. For a rotary-wing UAV with speed $V$, the propulsion power consumption can be expressed as \cite{zeng2019energy}:
\begin{align} \label{eq:energy}
	P(v) = P_0\left(1+\frac{3v^2}{U_{tip}^2}\right) & + P_{i}\left(\sqrt{1+\frac{v^4}{4v_0^4}}-\frac{v^2}{2v_0^2}\right)^{\frac{1}{2}} \\ & +
	\frac{1}{2}d_0 \psi d_Av^3,
\end{align}
where $P_0$ and $P_i$ are constants representing the blade profile power and induced power in hovering status, respectively. $P_i$ depends on the aircraft weight, air density $\psi$, and rotor disc area $d_A$, as specified in \cite{zeng2019energy}. Also, $U_{tip}$ denotes the tip speed of the rotor blade, $v_0$ is known as the mean rotor induced velocity in hovering, and $d_0$ and $s$ are the fuselage drag ratio and rotor solidity, respectively.

Therefore, with a given trajectory $q(t)$ where $q(t) \in R^2$ and $0 \leq t \leq T_m$, the propulsion energy consumption can be expressed as
\begin{equation}
	E(T_m,q(t))=\int_{0}^{T_m} P(||v(t)||)dt,
\end{equation}
where $||v(t)||$ is the instantaneous UAV speed.

\subsection{Simulation Results}
A $2\times2$ $\text{km}^2$ search area is considered to investigate the amount of time required to find the target. The algorithm is evaluated using two different probability distributions: Gaussian mixture and Gaussian-Uniform mixture distributions  \cite{gaussian-mixture}. A drone starts flying from the left-bottom corner of the search area. Table \ref{tab:timetocomplete} shows the average time to find the missing target for three different algorithms. In fact, the proposed algorithm is compared with two different algorithms: \textcolor{black}{Zigzag algorithm which is achieved via back and forth trajectories in the search area\cite{choset2001coverage}}. And, the algorithm used in \cite{lin2014hierarchical} in which it approximates the probability distribution map using Gaussian mixture model (GMM) and plans UAV motions heuristically using Gaussian probability density functions. As can be seen, the proposed algorithm is able to find the target in a shorter time in both distributions. Specifically, in case of Gaussian mixture distribution, it outperforms the Zigzag algorithm 5 times faster. 

\begin{table}[ht]
	\centering
	\caption{Average time (seconds) to find the missing target.}
	\begin{tabular}{c|c|c|}		
		\cline{2-3}
		&   \begin{tabular}{@{}c@{}} Gaussian mixture \\ distribution \end{tabular} & \begin{tabular}{@{}c@{}} Gaussian–Uniform mixture \\ distribution \end{tabular} \\ \hline
		\multicolumn{1}{|c|}{Zigzag algorithm} & 9800 & 10100 \\ \hline
		\multicolumn{1}{|c|}{\cite{lin2014hierarchical} } & 2160 & 11110 \\ \hline
		\multicolumn{1}{|c|}{Proposed algorithm} & 1960 & 8090 \\ \hline	
	\end{tabular}
	\label{tab:timetocomplete}
\end{table}

The energy performance of the proposed algorithm is evaluated and compared to two other algorithms as well. To this end, the power consumption model and parameters shown in Equation \eqref{eq:energy} are used for all cases and the efficiency is defined as the ratio of energy corresponding to benchmark algorithm to that of the proposed algorithm. 

\begin{table}[ht]
	\centering
	\caption{Energy consumption (kJ) to find the missing target.}
	\begin{tabular}{c|c|c|}		
		\cline{2-3}
		&   \begin{tabular}{@{}c@{}} Gaussian mixture \\ distribution \end{tabular} & \begin{tabular}{@{}c@{}} Gaussian–Uniform mixture \\ distribution \end{tabular} \\ \hline
		\multicolumn{1}{|c|}{Zigzag algorithm} & 2487 & 2564 \\ \hline
		\multicolumn{1}{|c|}{\cite{lin2014hierarchical} } & 564 & 2706 \\ \hline
		\multicolumn{1}{|c|}{Proposed algorithm} & 503 & 2132 \\ \hline	
	\end{tabular}
	\label{tab:energycpnsumption}
\end{table}

\begin{figure}[htb]	
	\centering
	\includegraphics[width=0.5\textwidth]{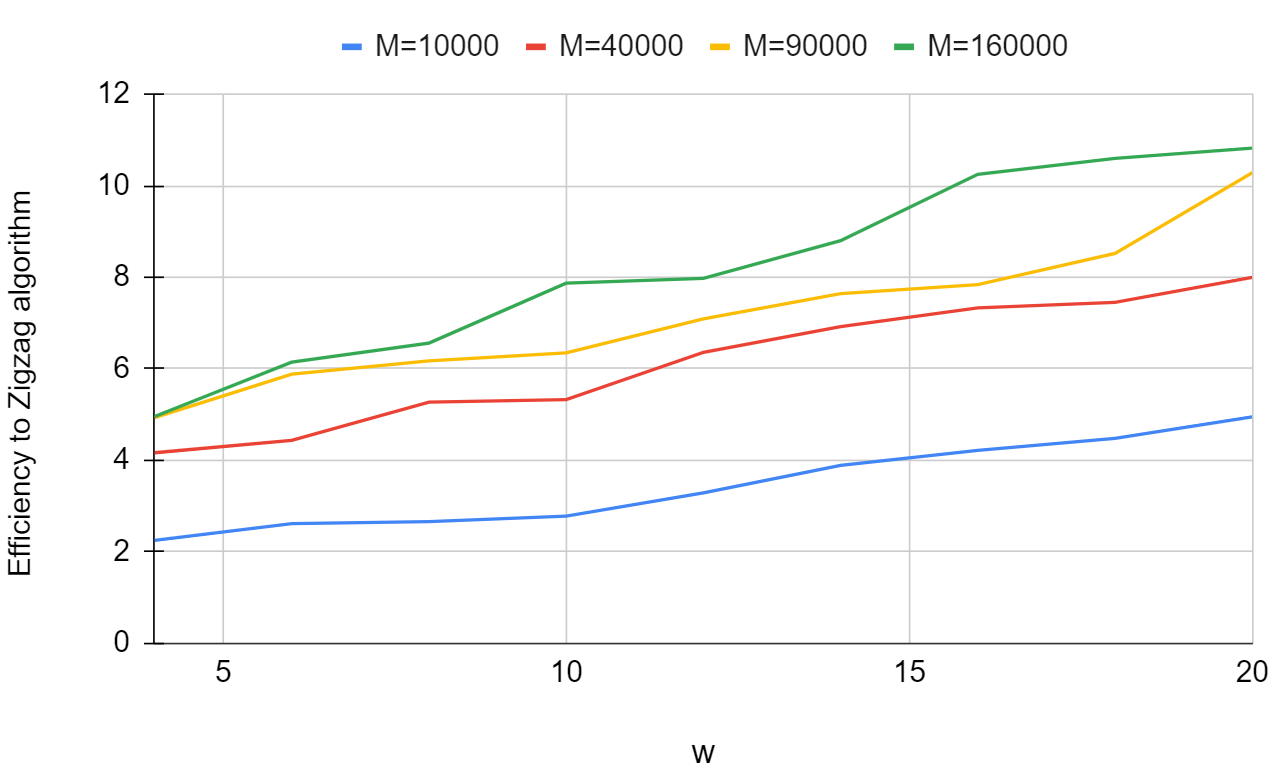}
	\caption{The effect of $W$ and $M$ in the path planning algorithm.}
	\label{fig.M_W_effects}
\end{figure}

Figure \ref{fig.M_W_effects} shows the effect of changing $W$ and $M$ in path planning algorithm where the location of the target follows a Gaussian mixture distribution. The vertical axis shows the performance ratio of the proposed algorithm to the Zigzag algorithm. As it can be seen, the performance ratio of the proposed algorithm with respect to the Zigzag algorithm improves with increasing $W$.  

\subsection{Experimental Results}
\begin{table}[t!]
	\centering
	\caption{\textcolor{black}{Parameters values in experimental evaluations.}}\textcolor{black}{
	\begin{tabular}{lll}
		\hline
		\multicolumn{1}{|c|}{\textbf{Parameter}}  &  \multicolumn{1}{c|}{\textbf{Value}} \\ \hline
		\multicolumn{1}{|c|}{Drone model}       &  \multicolumn{1}{c|}{DJI Mavic $2$ Zoom} \\ \hline
		\multicolumn{1}{|c|}{Target}            &  \multicolumn{1}{c|}{A firepit}\\ \hline
		\multicolumn{1}{|c|}{Target distribution}            &  \multicolumn{1}{c|}{Gaussian-Uniform}\\ \hline
		\multicolumn{1}{|c|}{$S$: Area}         &  \multicolumn{1}{c|}{234000SF} \\ \hline
		\multicolumn{1}{|c|}{$H$: Flight height}&  \multicolumn{1}{c|}{25 m} \\ \hline
		\multicolumn{1}{|c|}{$v$: Flight speed} &  \multicolumn{1}{c|}{20 $\frac{m}{s}$} \\ \hline 
	\end{tabular}}
	\label{tab:experimentValues}
\end{table}

 For the experimental results, extensive experiments have been conducted on the system to investigate end-to-end functionality, detection accuracy, and algorithm efficiency. In a real experiment, a scenario where the target is a fire spot is considered to be found within a large field of $234000 $SF. The location of the fire follows a Gaussian–Uniform mixture distribution. The drone flies at an altitude of $25$m from the ground with a speed of $20\frac{m}{s}$. The boundary of the flight is selected based on the search area. \textcolor{black}{A summary of experiments parameters are shown in Table \ref{tab:experimentValues}}. The instruments of the UAV Control Module are a phone and a UAV remote-control unit. With DJI Go App and DJI SDK, one can set the flight missions of the UAV and control its flight state. The searching path for the flight mission is generated using the probability map, and the drone flew to each point as expected. 
 
 It is required that the ground controller receives the detected target's information in a timely and convenient manner so that it could take the corresponding actions. Thus, the real-time transmission of the associated information is crucial. Real-time transmission is realized both in the UAV Control algorithm and the detection algorithm. With DJI Go App and DJI SDK, we can set the flight missions of the UAV and control its flight state. In addition, the real-time video can be displayed on the phone.
 
 At each point of operation, the images are captured and analyzed, then the probability map is updated based on the new observation. The searching algorithm finds the next optimal point based on the updated map. Table \ref{tab:exprimentresult} shows the average time to find the target and the corresponding energy consumption for our proposed algorithm and the Zigzag algorithm and the one proposed in \cite{lin2014hierarchical}. The average time to find the target was evaluated by measuring the amount of time spent to find the finding the target successfully. The energy consumption during the target detection mission was evaluated by measuring the UAV battery level during the mission and at the end of the mission.
 
 \begin{table}[ht]
 	\centering
 	\caption{Experimental results.}
 	\begin{tabular}{c|c|c|}		
 		\cline{2-3}
 		& Time to complete(S) & Energy consumption(kJ) \\ \hline
 		\multicolumn{1}{|c|}{Zigzag algorithm} & 1024 & 230.4 \\ \hline
 		\multicolumn{1}{|c|}{\cite{lin2014hierarchical} } & 1060 & 244.8 \\ \hline
 		\multicolumn{1}{|c|}{Proposed algorithm} & 920 & 213.4 \\ \hline	
 	\end{tabular}
 	\label{tab:exprimentresult}
 \end{table}


\section{Conclusion}
\label{sec:conclude}

In this paper, a general framework was proposed for the problem of object detection in which path planning and machine learning models are investigated. In particular, scenarios were considered in which objects such as a lost hiker in the snow or a fire spot in a given area are intended to be detected. Using a probability map, a path planning algorithm was proposed based on the Bayesian inference so that the shortest path leading to the object detection is designed. Along with the path planning algorithm, a residual neural network was utilized to capture the object. Since in the considered scenario the object is a person camouflaged in the snow or a fire spot in its initial levels, new dataset  was developed besides the existing ones, for training, validation, and testing of the neural network. The results are verified through simulation and experiment. It has been shown that the proposed method outperforms the existing ones in terms of average time spent to find the object and the energy consumption of the drone. \textcolor{black}{The present work provides a proof of concept for the SAR and fire detection operations. Therefore, there are several avenues for future work: One can apply other ML models such as ensemble ML models and investigate the existing trade-off between the detection performance and running time as well as approaches to improve this balance. Larger datasets are also of interest to be exploited to consolidate the model's generalization and robustness.} Furthermore, extending the proposed framework for a multi UAV setting is an interesting problem in an analysis and practical level. In this case, the complexity will increase as an efficient coordination among UAVs is required for an optimal operation. Another extension is to consider more constraints such as communication cost and scheduling to the UAVs operation especially in a multi UAV setting. Finally, using datasets for different situations such as images taken at night is another potential direction one can investigate.


\ifanonymous
{ }
\else

\fi



\section*{Acknowledgment}

The authors would like to thank Jarrett Austin of the Pelham Fire Department (Massachusetts) for all his assistance during the conducting of the experiments.
\bibliographystyle{IEEEtran}
\bibliography{bib/ref,bib/ref1,bib/refs1,bib/refssar,bib/relatedrefs}

\end{document}